# A LINEAR APPROXIMATION METHOD FOR PROBABILISTIC INFERENCE

Ross D. Shachter
Department of Engineering-Economic Systems
Stanford University, Stanford, CA 94305-4025
visiting at the Center for Health Policy Research and Education
125 Old Chemistry Building, Duke University, Durham, NC 27706

## Abstract

An approximation method is presented for probabilistic inference with continuous random variables. These problems can arise in many practical problems, in particular where there are "second order" probabilities. The approximation, based on the Gaussian influence diagram, iterates over linear approximations to the inference problem.

## Introduction

There have been a number of techniques developed in recent years for the efficient analysis of probabilistic inference problems, represented as Bayes' networks or influence diagrams [Lauritzen and Spiegelhalter 1988, Pearl 1986, Shachter 1988]. To varying degrees these methods exploit the conditional independence assumed and revealed in the problem structure to analyze problems in polynomial time, essentially polynomial in the number of variables and the size of the largest state space encountered during the evaluation. Unfortunately, there are many problems of interest for which the variables of interest are *continuous* rather than *discrete*, so the relevant state spaces become infinite and the polynomial complexity is of little help.

In this paper, an algorithm is presented which is based on a linear approximation to the problem structure. Each of the variables in the model is transformed, and the transformed variables are assumed to have a Gaussian joint distribution. Through successive iterations, this linear approximation is refined until it converges to a consistent solution. Although this method is an approximation rather than an analytical solution, it has proven quite accurate for a variety of problems in health technology assessment. It has tended to converge rapidly and, since each step is polynomial in the number of variables, this provides a polynomial heuristic for probabilistic inference with continuous variable.

The algorithm presented in this paper was motivated by a technique for medical technology assessment based on second order probabilities [Eddy 1988, Shachter et al 1987 1988]. The parameters of interest are the probabilities for different well-defined physical events. The probabilities are uncertain quantities and our prior knowledge about them is described by (usually "noninformative") probability distributions. The relevant medical evidence is then incorporated within a model to provide defensible, posterior distributions for these parameters.

There is an established philosophical basis for this approach, which provides a solid framework for knowledge acquisition in uncertain environments. Recent work argues persuasively that the established methodology for probabilistic reasoning applies *theoretically* to these second-order probabilities just as it does to the first-order kind [Kyburg 1987, Pearl 1987]. Nonetheless, the *practical* problems are considerable since the higher order probabilities are as a rule continuous distributions while the first order ones are usually discrete.

How then can these continuous probabilistic inference problems be analyzed? There are a several other approaches for dealing with this additional complexity besides the linear method.
1. **Conjugate priors:** If a model's structure allows it, prior distributions for parameters can be chosen from convenient families of conjugate distributions, so that the posterior distributions given the experimental evidence stay within those families [DeGroot 1972]. This is an analytical solution to continuous inference problem and the Gaussian model is one example of this approach. Unfortunately, conjugate families are not closed in general when experimental evidence bears (indirectly) on more than one basic parameter or for different forms of experimental evidence.
2. **Discretization:** Discrete techniques can be used by divide the sample space into intervals. However, processing time goes up with some power of the refinement, while resolution is only grows linearly with it. A similar approach to the one in this paper could be used to iterate, determining new discretizations after each solution.
3. **Numerical integration:** This is discretization of another sort. It is impractical for more than a few dimensions.
4. **Monte Carlo integration:** This is the state-of-the-art approach to numerical integration [Geweke 1988]. It can successfully solve the types of problems discussed here, without the distributional assumptions imposed by the linear approximation method. While it provides additional accuracy, it does so at substantially greater cost in computer time.

Although some of these other techniques might be more appropriate for a particular problem, the linear approximation possesses a unique combination of speed and generality, providing an efficient approximation to a large class of problems.

### Notation and Basic Framework

The original model consists of n variables, $Y_1, \ldots, Y_n$, represented by an influence diagram, a network with an acyclic directed graph. Each variable $Y_j$ corresponds to a node j in the diagram, thus the set $N = \{1, \ldots, n\}$ contains the nodes in the diagram. $Y_J$ denotes the set of variables corresponding to the indices J. Thus, the direct predecessors of node j, denoted by C(j), represent the set of conditioning variables $Y_{C(j)}$ for $Y_j$. When the order of the variables is significant the sequence **s**, a vector of indices, is used to represent the vector of variables, $Y_s$. A sequence s is called *ordered* if every predecessor for any node in the sequence precedes it in the sequence.

There are three types of variables represented in the influence diagram in this paper. *Basic parameters* are quantities for which a simple prior distribution is known. They have no conditioning variables, $C(j) = \emptyset$, and they are assumed to be mutually independent a priori. *Deterministic parameters* are quantities defined in terms of other parameters. They have conditioning variables but are assumed to be deterministic functions of those conditioning variables, so that their realizations would be known with certainty if the values of their conditioning variables were known. Finally, there are *experimental evidence* variables, whose realizations have been observed. They are characterized by a conditional distribution (a priori) and an observed value (a posteriori) and are assumed to have exactly one conditioning variable, either a basic or deterministic parameter. These variables and the assumptions about them are depicted in Figure 1. [For information, see Shachter et al 1988.]

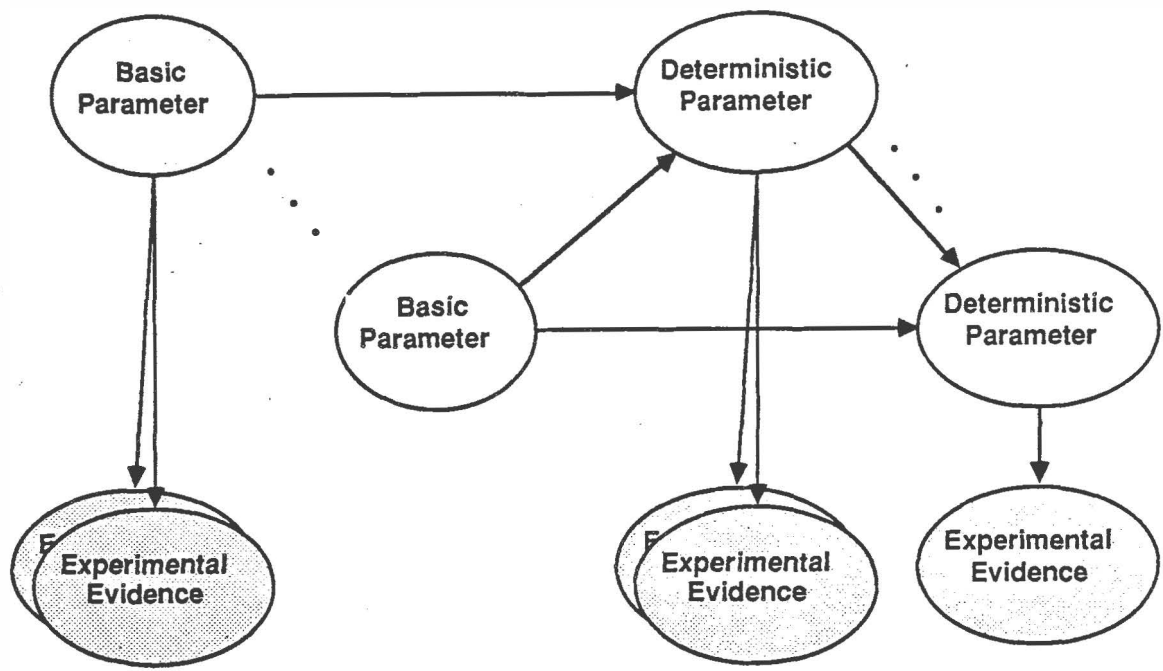

Figure 1. The model assumptions and the different types of variables

In the linear approximation method, a Gaussian variable, $X_j$, is associated with each parameter variable $Y_j$, by a deterministic function, $X_j = T_j(Y_j)$. The set of variables $X_N$ is assumed to have a multivariate normal joint distribution characterized by its means $E\,X_N$ and covariance matrix
$$\Sigma = \Sigma_{NN} = \text{Var}[X_N] = E[XX^T] - E[X]E[X^T].$$
Alternatively, the Gaussian influence diagram [Shachter and Kenley 1988] represents the multivariate Gaussian distribution through its conditional regression equations
$$X_j = E\,X_j + B_{C(j),j}^T [X_{C(j)} - E\,X_{C(j)}] + \varepsilon_j,$$
where $\varepsilon_j$ is a normal random variable with mean 0 and variance $v_j$, and B is a strictly upper triangular matrix of linear coefficients.

The resulting Gaussian model (if the original variables were integrated out) has the same structure as the original model, with basic and deterministic parameters and experimental evidence variables, except that they are assumed to have a multivariate normal distribution so that they can be manipulated using the operations of the Gaussian influence diagram. (Other similar techniques have been developed to exploit the Gaussian properties in a network representation [Pearl 1985]. Although the linear approximation method will be explained in terms of the Gaussian influence diagram, these other techniques could be used to implement it.)

One last bit of notation denotes the the revision of probabilities over time. The superscript $t$ as in $E^t\,X$ represents the prior expectation of X in the $t^{th}$ iteration and $E^t[X\mid D]$ represents its expectation after observing the experimental evidence. The superscript $^t$ will be omitted for readability whenever it is unambiguous to do so.

**Variable Transformations**

The fundamental property of the approach is that every variable in the model is transformed into a Gaussian variable, and the resulting multivariate Gaussian model will be maintained and manipulated, in order to provide indirect insight into the original variables and their dependence. Although the model could be embellished further, there are three basic transformations: *scaled*, *log-scaled* and *logistic-scaled*. These allow the representation of unbounded, semi-bounded, and bounded variables, respectively. Denoting a variable in the original model as Y, one in the transformed model as X, and the transformation function as T, the transformations are expressed in terms of *scaling parameters* a and b, where $a \neq b$:

1. **Scaled Transformation:** $Y \in (-\infty, +\infty)$
    $X = T(Y) = (Y - a)/(b - a)$.
    $T^{-1}(X) = a + (b - a)X$
    $T'(Y) = 1/(b - a)$

2. **Log-Scaled Transformation:** $Y \in (a, +\infty)$ if $a < b$ and $Y \in (-\infty, a)$ if $a > b$
    $X = T(Y) = \ln(Y - a)/(b - a)$
    $T^{-1}(X) = a + (b - a)e^X$
    $T'(Y) = 1/|b - a|$

3. **Logistic-Scaled Transformation:** $Y \in (a, b)$ if $a < b$ and $Y \in (b, a)$ if $a > b$
    $X = T(Y) = \ln(Y - a)/(b - Y)$
    $T^{-1}(X) = b + (a - b)/(1 + e^X)$
    $T'(Y) = 1/|Y - a| + 1/|Y - b|$

Of course, X and Y are random variables, so we must be able to transform from the *distribution* for X to the *distribution* for Y. We approximate this more complicated transformation by the function T, which maps the mean and variance of Y into the mean and variance for X, based on the distributional form for Y. These transformations would be exact if the $X_N$ were truly multivariate normal.

1. **Normal Distribution:** $(Y - a)/(b - a) \sim \text{Normal}(\mu, \sigma^2))$
    with $X = (Y - a)/(b - a)$ (scaled transformation)
    $(EX, \text{Var } X) = T(EY, \text{Var } Y) = ((EY - a)/(b - a), \text{Var } Y/(b - a)^2)$
    $(EY, \text{Var } Y) = T^{-1}(EX, \text{Var } X) = (a + (b - a)EX, (b - a)^2 \text{Var } X)$

2. **Lognormal Distribution:** $(Y - a)/(b - a) \sim \text{Lognormal}(\mu, \sigma^2)$
    with $X = \ln(Y - a)/(b - a)$ (log-scaled transformation)
    $(EX, \text{Var } X) = T(EY, \text{Var } Y) = (\mu, \sigma^2)$
        where $\sigma^2 = \text{Var } X = \ln[1 + \text{Var } Y (EY - a)^{-2}]$

302

and $\mu = EX = \ln[(EY - a)/(b - a)] - \sigma^2/2$.

$(EY, \text{Var } Y) = T^{-1}(EX, \text{Var } X)$
$= (a + (b - a) e^{\{EX + \text{Var } X/2\}}, (b - a)^2 (e^{\text{Var } X} - 1) e^{\{2EX + \text{Var } X\}})$.

3. **Beta Distribution:** $(Y - a)/(b - a) \sim \text{Beta}(\alpha, \beta)$

with $X = \ln(Y - a)/(b - Y)$ (logistic-scaled transformation)

$(EX, \text{Var } X) = T(EY, \text{Var } Y) = (\psi(\alpha) - \psi(\beta), \psi'(\alpha) + \psi'(\beta))$

where $\psi$ and $\psi'$ are the digamma and trigamma functions [Abramowitz and Stegun 1972],

$$\psi(z) = \frac{d \ln \Gamma(z)}{dz} = \frac{\Gamma'(z)}{\Gamma(z)} \approx \sum_{i=0}^{9} \frac{-1}{z+i} + \ln w - \frac{w^{-1}}{2} - \frac{w^{-2}}{12} + \frac{w^{-4}}{120} - \frac{w^{-6}}{252},$$

$$\psi'(z) = \frac{d^2 \ln \Gamma(z)}{dz^2} \approx \sum_{i=0}^{9} \frac{1}{(z+i)^2} + w^{-1} + \frac{w^{-2}}{2} + \frac{w^{-3}}{6} - \frac{w^{-5}}{30} + \frac{w^{-7}}{42} - \frac{w^{-9}}{30},$$

and $w = z + 10$.

There is no closed-form expression for the inverse function $T^{-1}$. If $\alpha$ and $\beta$ are large enough, then they can be approximated by

$\alpha_0 \approx .5 + (1 + e^{EX})/\text{Var } X$

and $\beta_0 \approx .5 + (1 + e^{-EX})/\text{Var } X$.

In general, however, $\alpha$ and $\beta$ can be estimated using Newton's method and curve fitting using the iterative formula:

$$\begin{bmatrix} \alpha_{k+1} \\ \beta_{k+1} \end{bmatrix} = \max \left\{ \begin{bmatrix} .5 \alpha_k \\ .5 \beta_k \end{bmatrix}, \begin{bmatrix} \alpha_k \\ \beta_k \end{bmatrix} - \begin{bmatrix} \psi'(\alpha_k) & -\psi'(\beta_k) \\ \psi''(\alpha_k) & \psi''(\beta_k) \end{bmatrix}^{-1} \begin{bmatrix} \psi(\alpha_k) - \psi(\beta_k) - E[X] \\ \psi'(\alpha_k) + \psi'(\beta_k) - \text{Var}[X] \end{bmatrix} \right\},$$

where $\psi''$ is the tetragamma function [Abramowitz and Stegun 1972],

$$\psi''(z) = \frac{d^3 \ln \Gamma(z)}{dz^3} \approx \sum_{i=0}^{9} \frac{-2}{(z+i)^3} - w^{-2} - w^{-3} - \frac{w^{-4}}{2} + \frac{w^{-6}}{6} - \frac{w^{-8}}{6}.$$

### Experimental Observations

The linear approximation requires that likelihood functions for experimental observations be derived in terms of the transformed, Gaussian parameter X on which the experimental evidence bears. Three kinds of experimental evidence are consider here, assuming samples from either a binomial or normal distribution. Of course, the method could be extended to other experimental designs.

1. **Normal experiment**, n exchangeable samples from Normal ($X, \sigma^2$) where $\sigma^2$ is known and the observed values have sample mean $m = \sum_j d_j / n$.

The likelihood function for the evidence is given by

$D \mid X \sim \text{Normal}(X, \sigma^2/n)$ with observation $d = m$.

2. **Normal experiment**, n > 2 exchangeable samples from Normal $(X, \sigma^2)$ where $\sigma^2$ is fixed but unknown and the observed values have sample mean m and sample variance $s = \sum_j (d_j - m)^2 / n$.

The likelihood is t-distributed but can be approximated by

$D \mid X \sim$ Normal $(X, s/(n-3))$ with observation $d = m$.

(Note: The preceding likelihoods can also be used for exchangeable samples from a lognormal distribution by transforming each sample.)

3. **Binomial experiment**, $D \sim$ Binomial $((Y-a)/(b-a), n)$ with s successes observed

with $X = \ln Y/(1-Y)$ (logistic-scaled transformation, $a = 0, b = 1$)

The likelihood is binomial-distributed but can be approximated by

$D \mid X \sim$ Normal $(X, v)$ with observation $d = v [x_2/v_2 - x_1/v_1]$

where $v = 1/[1/v_2 - 1/v_1]$,

$v_1 = \psi'(\alpha) + \psi'(\beta)$,

$v_2 = \psi'(\alpha + s) + \psi'(\beta + n - s)$,

$x_1 = \psi(\alpha) - \psi(\beta)$,

and $x_2 = \psi(\alpha + s) - \psi(\beta + n - s)$.

(The estimate is most accurate if $\alpha$ and $\beta$ are the prior parameters for Y. Alternatively, they can be set equal, to values such as .5 or 1, but they need not be.)

**Linear Approximation Algorithm**

1. The first step in the algorithm is to compute the linear approximations for each of the basic parameters and each of the experimental observations. These values will be used in each iteration of the algorithm. To estimate the original value for the remaining, deterministic parameters compute, in order,

$E^0 Y_j = E^0 [f_j (Y_{C(j)})] \approx f_j (E^0 Y_{C(j)})$

(approximating the expected value of the function by the function of the expected value) and set the conditional variance $Var^0 [Y_j \mid Y_{C(j)}] = 0$.

2. The iterative step in the algorithm proceeds until the algorithm has either converged or diverged. Define the relative difference from one iteration to the next as

$r_j^t = 0$ if $E^t [X_j \mid D] = E^{t-1} [X_j \mid D]$

$= \mid E^t [X_j \mid D] - E^{t-1} [X_j \mid D] \mid / \max \{ \mid E^t [X_j \mid D] \mid, \mid E^{t-1} [X_j \mid D] \mid \}$ otherwise.

Letting $r_{max}^t = \max_j \{ r_j^t \}$, *convergence* occurs when $r_{max}^t < \epsilon$ and *divergence* occurs when $r_{max}^t > r_{max}^{t-1} > ... > r_{max}^{t-m}$ for some m such as 3.

2a. The first step in each iteration is to compute the linear coefficient in the Gaussian influence diagram for the transformed variables. For each variable, in order, compute

$B_{ij}^t = \left[ \dfrac{\partial x_j}{\partial x_i} \right]_{x_N = E^{t-1}[X_N \mid D]}$,

taking advantage of the linearity of $X_N$. Now, using the approximation in 2d,

$$B_{ij}^t \approx \left[ \frac{\partial T_j(f_j(y_{C(j)}^{t-1}))}{\partial T_i(y_i^{t-1})} \right]_{y_N = E^{t-1}[Y_N | D]}$$

$$\approx \frac{T_j'(f_j(E^{t-1}[Y_{C(j)}|D])) \frac{\partial f_j}{\partial y_i}(E^{t-1}[Y_{C(j)}|D])}{T_i'(E^{t-1}[Y_i|D]])}$$

2b. For basic parameters and experimental outcomes, set the mean and conditional variances for the transformed variables to their original values. For each deterministic parameter, in order,

$$E^t X_j = E^t[T_j(Y_j)] = E^t[T_j(f_j(Y_{C(j)}))] \approx T_j(f_j(E^t Y_{C(j)}))$$

$$\approx T_j(f_j(E^{t-1}[Y_{C(j)}|D])) + \sum_{i \in C(j)} B_{ij}^t (E^t[X_i] - E^{t-1}[X_i|D])$$

and set the conditional variance to zero. (This is the first order approximation to E X, relative to the posterior from the previous iteration, in the same spirit as Maybeck [1982].)
Afterwards, compute the unconditional variance $\Sigma$, (assuming the variables are ordered),

for $j = 1, \ldots, n$:
  let $s = (1, \ldots, j-1)$
  $\Sigma_{sj} = \Sigma_{js}^T = \Sigma_{ss} B_{sj}$
  $\Sigma_{jj} = v_j + B_{sj}^T \Sigma_{ss} B_{sj}$.

2c. The evidence must now be instantiated. This can be performed in several ways, but the theoretical process is represented by the two matrix equations:

$$E[X_N | D] = E X_N + \Sigma_{ND} \Sigma_{DD}^{-1} (d - E D)$$

and $\quad Var[X_N | D] = \Sigma_{NN} - \Sigma_{ND} \Sigma_{DD}^{-1} \Sigma_{DN}$.

2d. Finally, compute the estimated posterior value for each basic and deterministic parameter in the model, using the inverse transform approximation,

$$(E^t[Y_j | D], Var^t[Y_j | D]) \approx T_j^{-1}(E^t[X_j | D], Var^t[X_j | D]).$$

## Conclusions

The method presented here provides a simple, efficient framework for approximating probabilistic inference over continuous distributions. The empirical evidence with the procedure has shown it to be fairly accurate and fast when there is sufficient data. (It can have convergence problems when the priors are flat and there is little experimental evidence.)

Some simple changes can improve the accuracy of the method. First, multiple (conditionally independent) experimental evidence for the same parameter can be "pooled" into a single experiment for the purposes of the approximation. Second, deterministic relationships which are

305

analytically linear can be recognized symbolically, and the corresponding regression coefficients computed in advance. These include linear combinations of scaled variables, products of log-scaled variables, and odds-ratios of logistic-scaled variables.

Finally, there is a useful byproduct of the linear approximation algorithm: an estimate of the correlation between any two of the model parameters,

$$\text{Corr}[X_i, X_j | D] = 0 \quad \text{if } \text{Var}[X_i | D] = 0 \text{ or } \text{Var}[X_j | D] = 0$$
$$= \text{Cov}[X_i, X_j | D] \, (\text{Var}[X_i | D] \, \text{Var}[X_j | D])^{-1/2} \quad \text{otherwise}.$$

This provides insight into the sensitivity of the posterior estimates to changes in prior distributions or additional experimental evidence.

### Acknowledgements

This research was supported by the John A. Hartford Foundation and the National Center for Health Services Research under Grants 1R01-HS-05531-01 and 5R01-HS-05531-02. I thank my colleagues David Eddy, Vic Hasselblad, and Robert Wolpert for their encouragement and collaboration.